# Exploring the Determinants of Pedestrian Crash Severity Using an AutoML Approach

**Amir Rafe [1*], Patrick A. Singleton [2]**

[1] Utah State University, Department of Civil and Environmental Engineering, 4110 Old Main Hill, Logan, UT 84322-4110, USA. Email: amir.rafe@usu.edu.
[2] Utah State University, Department of Civil and Environmental Engineering, 4110 Old Main Hill, Logan, UT 84322-4110, USA. Email: patrick.singleton@usu.edu.

## ABSTRACT

This study investigates pedestrian crash severity through Automated Machine Learning (AutoML), offering a streamlined and accessible method for analyzing critical factors. Utilizing a detailed dataset from Utah spanning 2010-2021, the research employs AutoML to assess the effects of various explanatory variables on crash outcomes. The study incorporates SHAP (SHapley Additive exPlanations) to interpret the contributions of individual features in the predictive model, enhancing the understanding of influential factors such as lighting conditions, road type, and weather on pedestrian crash severity. Emphasizing the efficiency and democratization of data-driven methodologies, the paper discusses the benefits of using AutoML in traffic safety analysis. This integration of AutoML with SHAP analysis not only bolsters predictive accuracy but also improves interpretability, offering critical insights into effective pedestrian safety measures. The findings highlight the potential of this approach in advancing the analysis of pedestrian crash severity.

## INTRODUCTION

Pedestrian safety is a critical concern in urban transportation networks, as pedestrians represent a highly vulnerable group in traffic-related incidents. The urgency to enhance their protection and mitigate the severity of pedestrian-involved crashes cannot be overstated. Recent statistics from the National Highway Traffic Safety Administration (NHTSA) (2022) highlight this concern: in 2021, there were 7,388 pedestrian fatalities in the United States due to traffic accidents. The nature and severity of pedestrian crashes are influenced by a multitude of factors. Analysis of NHTSA data from 2021 reveals that pedestrian fatalities are predominantly occurring in urban areas (83%), at intersections (23%), and in conditions of low visibility, such as under dark lighting (39%). Furthermore, a report from the World Health Organization (WHO, 2022) underscores a startling fact: the likelihood of fatality for a pedestrian struck by a vehicle increases exponentially with the vehicle's speed, rising by 4.5 times with every 10mph increase in speed. Understanding these contributing elements is vital in devising strategies to decrease the number of pedestrian casualties and enhance overall traffic safety for pedestrians.

To effectively discern the determinants of pedestrian injury severity, it is common practice to examine historical crash data and employ predictive modeling techniques. The field of pedestrian safety research extensively utilizes a variety of analytical methods, including statistical analysis, machine learning, and deep learning algorithms, to assess the impact of different factors on the severity of pedestrian-involved accidents. Enhancing the precision and efficiency of these predictive models is crucial for policymakers and traffic safety engineers. A thorough understanding of the key variables that contribute to reducing pedestrian injuries and fatalities enables these stakeholders to focus on the most effective safety interventions in their planning. Furthermore, the efficacy of these interventions can be assessed through comparative studies conducted before and after their implementation. In summary, refining these predictive models is instrumental in developing focused and efficacious strategies to protect pedestrians and decrease the incidence of injuries and fatalities in traffic crashes.

The advancement of artificial intelligence and the integration of machine learning (ML) in various fields have led to this study's objective: **introducing an Automated Machine Learning (AutoML) methodology for predicting the severity of pedestrian crashes.** This AutoML approach is designed to simplify the process of evaluating different ML methods on safety data, selecting the most effective one. It particularly focuses on providing an accessible, low-code solution for transportation engineers and decision-makers who may not be well-versed in computer science yet seek high-accuracy predictions to inform reliable decision-making. Furthermore, this study aims to pinpoint key factors influencing pedestrian injury severities. To accomplish this, we utilized pedestrian crash data from Utah spanning the years 2010 to 2021. We applied various ML methods and developed a web application centered around AutoML, making it adaptable to any crash severity dataset. Additionally, to interpret the results from the chosen ML method, we utilized SHapley Additive exPlanations (SHAP), a method grounded in game theory.

In the sections that follow, we will conduct a literature review to examine prior studies that have implemented innovative techniques in crash severity prediction. This will be followed by a detailed presentation of our data and methodologies, the results we obtained, and a discussion focusing on the principal findings, interpretation of the model, and our concluding insights.

## LITERATURE REVIEW

Extensive research has been conducted to understand the multifaceted nature of pedestrian crash severity, with studies examining a range of influential factors. A thorough examination of the current literature, as summarized by (Shrinivas et al., 2023), identifies several key determinants. These include demographic aspects of the pedestrian such as age and gender, characteristics of the vehicle involved including its speed and type, environmental factors like the location of the collision and the time it occurred, and behavioral elements such as the involvement of alcohol or drugs in either the pedestrian or the driver. Additionally, the use of safety gear, such as helmets and reflective clothing, has been noted as an important factor.

Various studies have implemented ML approaches to assess pedestrian crash severity. Al-Mistarehi et al. (2022) explored several techniques, including decision trees, k-nearest neighbors (KNN), naive Bayes, and AdaBoost, and found that the random forest model outperformed others in accurately predicting different injury types, boasting the lowest error rate. In a different study, Goswamy et al. (2023) employed XGBoost and Random Parameters Discrete Outcome Models (RPDOM), uncovering that Rectangular Rapid Flashing Beacons (RRFB) notably reduce the severity of nighttime pedestrian crashes. Their study highlighted XGBoost's superior predictive accuracy of 97%, in contrast to RPDOM's 73.8%. Rahman et al.'s (2023) research on Utah state highways utilized boosted decision trees to identify risk factors affecting pedestrian crash frequencies. Their analysis revealed a correlation between higher motor vehicle and pedestrian volumes, transit stops, and certain demographic areas with increased crashes, while also noting a 'safety-in-numbers' effect where higher pedestrian volumes corresponded to lower crash rates. Effati & Vahedi Saheli (2022) compared logistic regression and classification and regression trees (CART) for estimating pedestrian crash occurrences in rural settings. Their findings suggested a marginally better performance of logistic regression over CART in identifying risky segments for pedestrians. Lastly, Yang et al. (2022) investigated the effectiveness of three ML methods—Support Vector Machines (SVM), Ensemble Decision Trees (EDT), and k-Nearest Neighbors (KNN)—each optimized using a Bayesian algorithm for predicting pedestrian fatalities in road crashes. The study found that while all models improved in performance due to optimization, the KNN model showed the most significant accuracy enhancement. However, the SVM and EDT models still exhibited higher overall accuracy than the KNN model.

Following the exploration of various machine learning models in crash severity analysis, it is pertinent to discuss the benefits and challenges these models present. One significant advantage is their higher predictive accuracy. Machine learning methods can outperform traditional statistical models, especially in situations involving complex, non-linear relationships among variables or in the analysis of large and intricate datasets (Komol et al., 2021). Additionally, these models excel at identifying and understanding the importance of different features (explanatory variables), elucidating intricate relationships with crash severity. This capability often yields insights that might be less apparent or more difficult to extract using statistical models (Komol et al., 2021). However, machine learning models are not without their limitations. A notable challenge is their interpretability; these models can be more complex and less intuitive than their statistical counterparts, potentially complicating the explanation of how various factors relate to crash severity (Infante et al., 2022). Moreover, there is the risk of overfitting—where a model becomes too tailored to the training data, losing its ability to generalize and perform accurately on new, unseen data. This issue is particularly prevalent when models are trained with a vast number of features or a limited dataset (Komol et al., 2021).

Addressing the limitations of traditional machine learning approaches, AutoML stands as a promising solution in crash severity analysis. AutoML enhances efficiency by simplifying the model selection process, reducing the need for in-depth machine learning expertise and computational resources (Angarita-Zapata et al., 2021). This is particularly beneficial in complex scenarios like crash severity prediction, where choosing an appropriate model can be challenging. Moreover, AutoML can test a broader range of ML methods on the data, unlike traditional studies that often explore a limited selection. Empirical results indicate that AutoML's capabilities go beyond mere convenience; it can match or even exceed the performance of conventional methods with reduced human intervention (Angarita-Zapata et al., 2021). This reflects AutoML's sophisticated algorithmic prowess, adept at sifting through various machine learning models to identify the most effective one for a specific dataset. The implications for research in road safety systems are profound. AutoML holds the potential to refine model selection and enhance prediction accuracy in this vital field. Recognizing these encouraging outcomes, this paper suggests a broader investigation into the utility of AutoML for pedestrian crash severity analysis. Such exploration could yield deeper insights into its efficacy and versatility, paving the way for more sophisticated and precise road safety systems.

## DATA AND METHOD

### DATA AND VARIABLES

For our research, we employed crash data reported by the Utah Department of Public Safety (UDPS, 2023), focusing on pedestrian crash severities in Utah from 2010 to 2021. The severity of pedestrian crashes in this study was categorized into three levels: fatal, serious injury, and minor injury. Figure 1 illustrates the geographical distribution of these crashes, including a heatmap that specifically emphasizes the locations of fatal crashes within the dataset. This figure highlights that the density of pedestrian crashes is predominantly concentrated in Salt Lake City, followed by Ogden and Provo.

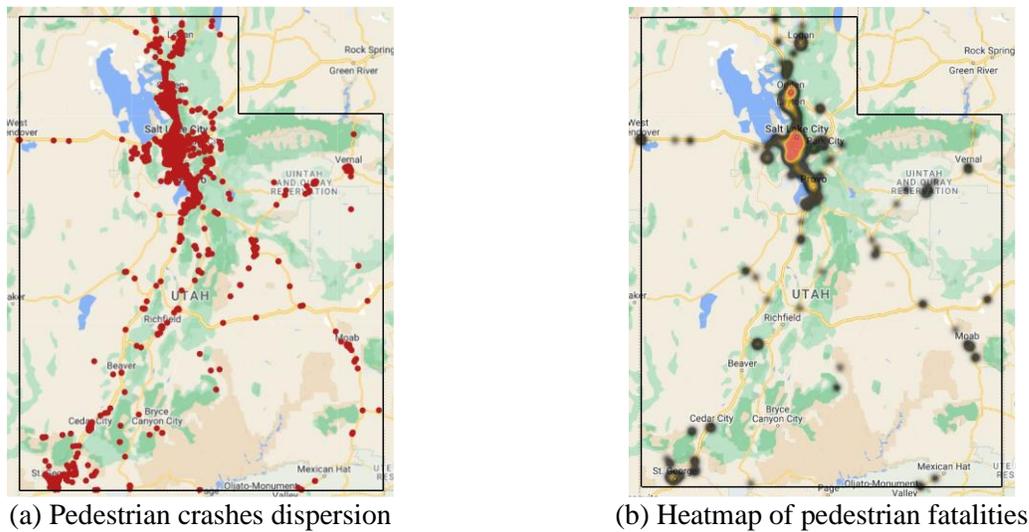

(a) Pedestrian crashes dispersion    (b) Heatmap of pedestrian fatalities

**Figure 1 The spatial configuration of pedestrian crashes**

In this study, we analyzed 8,319 pedestrian crashes and examined the influence of 17 explanatory variables on crash severity. Table 1 presents the descriptive statistics for these variables. For the scope of this study, non-injury pedestrian crashes (Property Damage Only, PDO) were not included, as the primary focus was on analyzing factors influencing the severity of injuries in pedestrian crashes. Regarding crash severity categorization, the 'minor injury' level in this research encompasses both 'minor injury' and 'possible injury' classifications according to the KABCO scale. According to the data presented in Table 1, males are more frequently involved in pedestrian crashes, representing 59% of all incidents, and they account for a higher proportion of fatal (65%) and serious injuries (62%). The age distribution presents a noteworthy finding: individuals aged 30 to 59 years are involved in 34% of crashes and constitute 42% of fatalities, while the younger cohort of 10 to 29 years old exhibits a higher incidence of serious injuries, at 41%. Substance influence is also a critical variable; crashes with both alcohol and drugs detected are less common but have a 100% fatality rate, highlighting the exacerbated risk of pedestrian fatalities when both substances are involved. Most pedestrian crashes happened under conditions where DUI was not reported (97%) and without distracted driving (92%).

The involvement of teenage drivers appears in 10% of fatal cases, suggesting a potential area for intervention. Intersections emerged as hotspots for pedestrian crashes, comprising 61% of all incidents. Concerningly, close to 3% of these intersection-related crashes had fatal outcomes for pedestrians. As for the nature of the crashes, incidents involving both left-turn and right-turn movements showed comparable rates of occurrence. Crashes predominantly occur in urban settings (97%) and on arterial roads (52%), pinpointing these as high-risk environments. The data also reveals that a significant number of fatal (37%) occur in dark-not-lighted conditions, which may signal a need for improved lighting to enhance pedestrian safety.

For uniform encoding of the dataset, numerical assignments were utilized such as "Yes" being coded as 1, "No" as 0; "Male" as 1, "Female" as 0; "Rural" as 1, "Urban" as 0. Other categories were numerically encoded in ascending order, beginning with 1 as outlined in Table 1.

Moreover, age was treated as a continuous variable, rather than categorizing it into discrete age groups.

**Table 1. Descriptive statistics of the variables**

| Characteristics | Class | Total | Fatal | Serious injury | Minor injury |
|---|---|---|---|---|---|
| Pedestrian crashes | | 8319 | 476 (6%) | 1363 (16%) | 6480 (78%) |
| Sex | Male | 4933 (59%) | 309 (65%) | 849 (62%) | 3775 (58%) |
| | Female | 3386 (41%) | 167 (35%) | 514 (38%) | 2705 (42%) |
| Age group | 0 to 9 | 782 (9%) | 33 (7%) | 118 (9%) | 631 (10%) |
| | 10 to 29 | 3795 (46%) | 119 (25%) | 556 (41%) | 3120 (48%) |
| | 30 to 59 | 2842 (34%) | 202 (42%) | 509 (37%) | 2131 (33%) |
| | > 59 | 900 (11%) | 122 (26%) | 180 (13%) | 598 (9%) |
| Alcohol-drug test result | Both-Positive | 11 (0%) | 11 (2%) | 0 (0%) | 0 (0%) |
| | Drug-Positive | 34 (0%) | 34 (7%) | 0 (0%) | 0 (0%) |
| | Alcohol-Positive | 15 (0%) | 13 (3%) | 2 (0%) | 0 (0%) |
| | Negative | 9 (0%) | 9 (2%) | 0 (0%) | 0 (0%) |
| | Not related | 8250 (99%) | 409 (86%) | 1361 (100%) | 6480 (100%) |
| DUI | No | 8102 (97%) | 413 (87%) | 1305 (96%) | 6384 (99%) |
| | Yes | 217 (3%) | 63 (13%) | 58 (4%) | 96 (1%) |
| Distracted driving | No | 7651 (92%) | 430 (90%) | 1218 (89%) | 6003 (93%) |
| | Yes | 668 (8%) | 46 (10%) | 145 (11%) | 477 (7%) |
| Teenage driver involved | No | 7523 (90%) | 432 (91%) | 1205 (88%) | 5886 (91%) |
| | Yes | 796 (10%) | 44 (9%) | 158 (12%) | 594 (9%) |
| Holiday | No | 7308 (88%) | 397 (83%) | 1185 (87%) | 5726 (88%) |
| | Yes | 1011 (12%) | 79 (17%) | 178 (13%) | 754 (12%) |
| Right-turn involved | No | 6743 (81%) | 464 (97%) | 1254 (92%) | 5025 (78%) |
| | Yes | 1576 (19%) | 12 (3%) | 109 (8%) | 1455 (22%) |
| Intersection involved | Yes | 5091 (61%) | 136 (29%) | 718 (53%) | 4237 (65%) |
| | No | 3228 (39%) | 340 (71%) | 645 (47%) | 2243 (35%) |
| Left-turn involved | No | 6652 (80%) | 441 (93%) | 1144 (84%) | 5067 (78%) |
| | Yes | 1667 (20%) | 35 (7%) | 219 (16%) | 1413 (22%) |
| Work zone involved | No | 7953 (96%) | 447 (94%) | 1298 (95%) | 6208 (96%) |
| | Yes | 366 (4%) | 29 (6%) | 65 (5%) | 272 (4%) |
| Road type | Urban | 8082 (97%) | 419 (88%) | 1296 (95%) | 6367 (98%) |
| | Rural | 237 (3%) | 57 (12%) | 67 (5%) | 113 (2%) |
| Functional class | Local | 2526 (30%) | 71 (15%) | 352 (26%) | 2103 (32%) |
| | Collector | 1493 (18%) | 71 (15%) | 232 (17%) | 1190 (18%) |
| | Arterial | 4300 (52%) | 334 (70%) | 779 (57%) | 3187 (49%) |
| Roadway surface is dry | Yes | 7175 (86%) | 409 (86%) | 1181 (87%) | 5585 (86%) |
| | No | 1144 (14%) | 67 (14%) | 182 (13%) | 895 (14%) |
| Lighting condition | Dark-Not lighted | 1117 (13%) | 176 (37%) | 285 (21%) | 656 (10%) |
| | Dark-Lighted | 1836 (22%) | 138 (29%) | 332 (24%) | 1366 (21%) |
| | Daylight | 4953 (60%) | 141 (30%) | 678 (50%) | 4134 (64%) |
| | Dusk | 224 (3%) | 10 (2%) | 40 (3%) | 174 (3%) |
| | Dawn | 189 (2%) | 11 (2%) | 28 (2%) | 150 (2%) |
| Weather condition | Clear | 6378 (77%) | 355 (75%) | 1068 (78%) | 4955 (76%) |
| | Cloudy | 1141 (14%) | 69 (14%) | 176 (13%) | 896 (14%) |
| | Rain | 484 (6%) | 31 (7%) | 80 (6%) | 373 (6%) |
| | Fog, Smog | 24 (0%) | 3 (1%) | 4 (0%) | 17 (0%) |
| | Snowing | 201 (2%) | 10 (2%) | 26 (2%) | 165 (3%) |
| | Others | 91 (1%) | 8 (2%) | 9 (1%) | 74 (1%) |
| Vertical alignment | Level | 6515 (78%) | 360 (76%) | 1108 (81%) | 5047 (78%) |
| | Uphill | 57 (1%) | 3 (1%) | 7 (1%) | 47 (1%) |
| | Downhill | 50 (1%) | 2 (0%) | 12 (1%) | 36 (1%) |
| | Others | 1697 (20%) | 111 (23%) | 236 (17%) | 1350 (21%) |

**METHOD**

As previously mentioned, this study utilized an AutoML methodology to investigate the impact of various explanatory variables (EV), or features, on the outcomes of pedestrian crash injuries, based on the pipeline depicted in Figure 2. This methodology encompasses several critical steps, each employing specific techniques to ensure the robustness and accuracy of the predictive models.

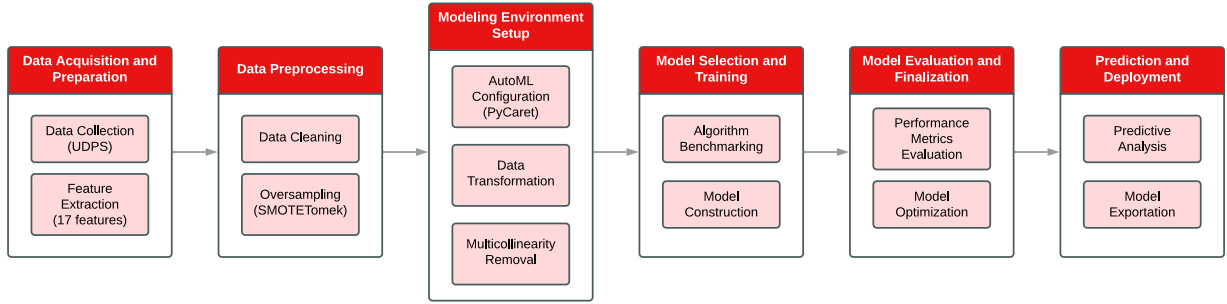

**Figure 2. The AutoML pipeline for pedestrian crash severity analysis in this study**

Initially, the study embarked on data acquisition, sourcing information from the UDPS and conducting feature extraction to identify the most relevant variables for the analysis. Subsequent data preprocessing involved cleaning and addressing class imbalance through oversampling methods to prepare a reliable dataset for model training. To address class imbalance in the dataset, the Synthetic Minority Over-sampling Technique coupled with Tomek link removal (SMOTETomek) (Batista et al., 2003) was employed to balance the severity classes. This method not only augments the dataset with additional synthetic minority class examples but also refines the class boundaries, thereby enhancing the model's predictive accuracy for less represented classes.

In setting up the AutoML environment, PyCaret (Moez Ali, 2020) was instrumental in configuring the necessary parameters, including data normalization and transformation processes. Additionally, features exhibiting high multicollinearity were identified and removed to prevent potential overfitting and improve model interpretability. The core of the AutoML approach was the model selection and training phase. Here, a variety of machine learning algorithms were evaluated and benchmarked to identify the model that best fits the data, considering the complex nature of crash severity outcomes. This was followed by a comprehensive evaluation of the chosen model, utilizing a suite of performance metrics such as F1 score, accuracy, precision, and recall assessing its predictive capabilities. Accuracy is the metric that measures the percentage of correctly classified observations across all categories. Precision quantifies the proportion of true positives among all positive predictions, indicating the model's exactness. Recall measures the proportion of actual positives that were correctly identified, reflecting the model's completeness. The F1 score, a calculated harmonic mean of precision and recall, provides a balanced measure of the model's precision and robustness in identifying positive classes. The mathematical formulations for these metrics are delineated by Equations 1 to 4.

$$accuracy = \frac{True\ Positive + True\ Negative}{True\ Positive + True\ Negative + False\ Positive + False\ Negative} \quad (1)$$

$$precision = \frac{True\ Positive}{True\ Positive + False\ Positive} \quad (2)$$

$$recall = \frac{True\ Positive}{True\ Positive + False\ Negative} \quad (3)$$

$$F1\ score = \frac{2 \times (precision \times recall)}{precision + recall} \quad (4)$$

Finally, the selected model underwent optimization and finalization, which involved fine-tuning to enhance its performance. The robustness of the model was validated, and upon satisfactory evaluation, it was deployed for predictive analysis.

In our study, we employed SHAP (Lundberg et al., 2018; Lundberg & Lee, 2017) to interpret the outcomes of selected ML model. SHAP provides a detailed measure of feature importance, assigning an importance value to each EV for a given prediction. This approach enhances the transparency of model predictions by translating them in terms of input EVs. Originating from cooperative game theory, SHAP values distribute the "payout" (in this case, the predicted crash severity) among the "players" (the EVs), based on their individual contributions to the prediction. To understand this mathematically, if f(x) represents the model's prediction for a specific instance x, and $E[f(X)]$ represents the expected model prediction (calculated as the mean prediction over the training set), then the contribution of each EV can be quantitatively expressed using an additive attribution model.

$$f(x) - E[f(X)] = \sum_{i=1}^{N} \varphi_i \quad (5)$$

Additionally, the significance of each EV, also known as the Shapley value for the i-th EV, denoted as $\varphi_i$ is computed as follows:

$$\varphi_i = \sum_{S \subseteq N\{i\}} \left[ \frac{|S|!(|N|-|S|-1)!}{|N|!} \right] (f_i(S \cup \{i\}) - f_i(s)) \quad (6)$$

In this context, $N$ is the set of all EVs, $S$ is a subset of $N$ that includes the i-th EV, |S| is the size of S, and $f_i$ is a version of f where only the EVs in $S$ and $i$ (if it's included) are used.

## MODEL RESULTS

The evaluation of various machine learning models yielded a range of performance outcomes, as summarized in Table 2. This table provides a detailed comparison of the models based on key performance metrics. The Random Forest model stood out in our comparative analysis of machine learning algorithms for predicting pedestrian crash severity. To assess its predictive performance, we employed several metrics visualized in Figures 3. The Receiver Operating Characteristic (ROC) curve and its corresponding Area Under the Curve (AUC) value in Figure 3 (d) measure the model's ability to correctly classify the severity levels, with a higher AUC reflecting a better overall performance. The confusion matrix in Figure 3 (c) provides a detailed breakdown of the model's predictions, showing the proportions of true positives, false positives, true negatives, and false negatives, which are critical for understanding the model's classification accuracy. In Figure 3 (b), the Precision-Recall Curve offers a view of the model's precision, or the accuracy of predicting positive classes, against its recall, the model's ability to capture all actual positives, which is particularly important in the context of imbalanced classes. Finally, Figure 3 (a), through SHAP values, ranks the features by their importance, showing how each one influences the model's predictions and providing insights into the decision-making process of the Random Forest model. These metrics collectively paint a comprehensive picture of the model's performance and its interpretive power regarding the factors that affect pedestrian crash severity.

**Table 2. Performance metrics of ML models for pedestrian crash severity prediction**

| Model | Accuracy | AUC | Recall | Precision | F1 score |
|---|---|---|---|---|---|
| Random Forest Classifier | 0.83 | 0.94 | 0.83 | 0.83 | 0.83 |
| Extra Trees Classifier | 0.82 | 0.93 | 0.82 | 0.82 | 0.82 |
| Extreme Gradient Boosting | 0.81 | 0.93 | 0.81 | 0.81 | 0.81 |
| Decision Tree Classifier | 0.80 | 0.87 | 0.80 | 0.80 | 0.80 |
| CatBoost Classifier | 0.80 | 0.93 | 0.80 | 0.80 | 0.80 |
| Light Gradient Boosting Machine | 0.78 | 0.92 | 0.78 | 0.78 | 0.78 |
| K Neighbors Classifier | 0.75 | 0.89 | 0.75 | 0.75 | 0.75 |
| Gradient Boosting Classifier | 0.69 | 0.87 | 0.69 | 0.70 | 0.69 |
| Ada Boost Classifier | 0.66 | 0.80 | 0.66 | 0.66 | 0.66 |
| Logistic Regression | 0.60 | 0.79 | 0.60 | 0.60 | 0.60 |
| Linear Discriminant Analysis | 0.59 | 0.79 | 0.59 | 0.61 | 0.60 |
| Ridge Classifier | 0.60 | 0.00 | 0.60 | 0.60 | 0.60 |
| SVM - Linear Kernel | 0.58 | 0.00 | 0.58 | 0.57 | 0.55 |
| Quadratic Discriminant Analysis | 0.57 | 0.77 | 0.57 | 0.57 | 0.55 |
| Naive Bayes | 0.57 | 0.77 | 0.57 | 0.58 | 0.55 |
| Dummy Classifier | 0.33 | 0.50 | 0.33 | 0.11 | 0.16 |

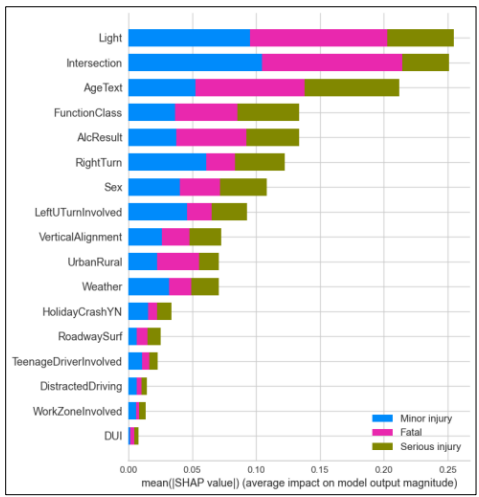

(a) Feature importance Derived from SHAP Values

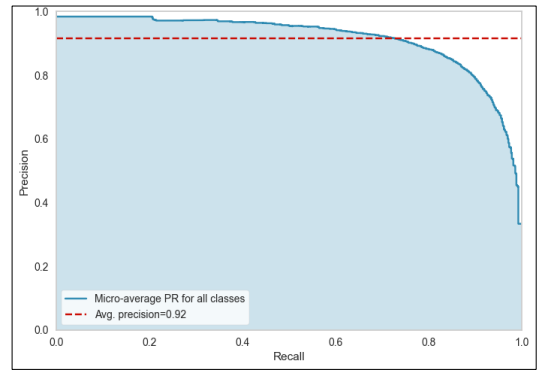

(b) Precision-Recall curve

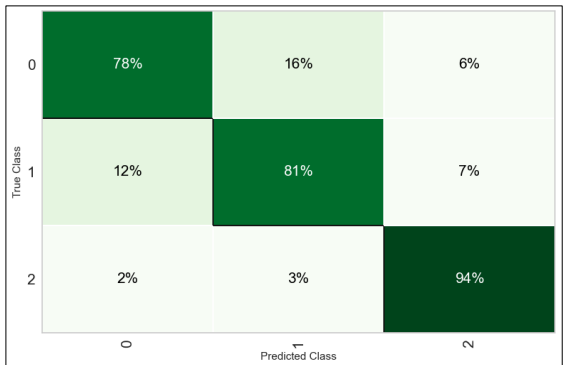

(c) Confusion matrix
(0: Minor injury, 1: Serious injury, 2: Fatal)

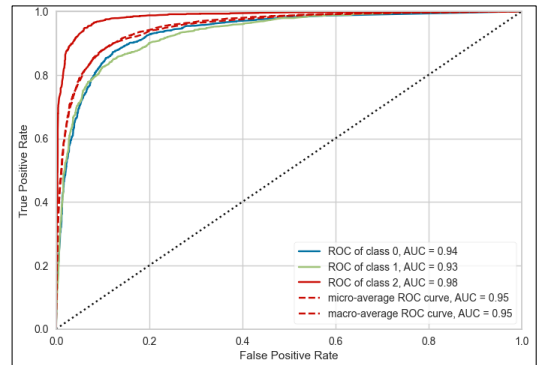

(d) ROC curves

**Figure 3. Comprehensive evaluation results of the Random Forest model**

## MODEL INTERPRETATION AND DISCUSSION

In this section, we delve into the results yielded by the AutoML process, with a particular focus on the Random Forest classifier, which was identified as the best-performing model. The Random Forest model demonstrated superior performance with an AUC of 0.94, indicating a high capability to distinguish between the different severity levels of pedestrian crashes. The ROC curve, which plots the true positive rate against the false positive rate, further confirmed the model's robustness in classification across varying thresholds. The model's precision and recall were balanced, as evidenced by the Precision-Recall Curve, with an average precision of 0.92. This balance is crucial in practice, as it ensures that the model is not only accurate in its predictions but also consistent in identifying the true instances of each class.

The confusion matrix provided a more granular view of the model's predictive accuracy, revealing that the model successfully classified 78% of the class 0 (minor injury) instances, 81% of class 1 (serious injury), and 94% of class 2 (Fatal) instances correctly. This distribution indicates a strong predictive power, particularly for the severe class, which is often the most challenging to predict due to the typically lower number of instances.

The SHAP analysis, which offers a game-theoretic approach to feature importance, shed light on the impact of various explanatory variables on the model's predictions. Features like lighting condition, whether the crash occurred at the intersection, and age were shown to have a substantial influence on the model's output, indicating that these factors play a significant role in the severity of pedestrian crashes. The SHAP values allow us to quantify the contribution of each feature to the prediction of each class, providing a clear picture of the underlying factors that the model considers when making predictions.

In assessing the Random Forest model's interpretation for each injury severity class, the SHAP value plots reveal the varied influence of explanatory variables. Figure 4 displays a SHAP summary plot that depicts how EV features influence the likelihood of each crash severity class. Each feature is represented by a row in the plot, with dots colored according to the feature's value—red indicating higher values and blue for lower ones. The placement of the dots horizontally indicates the direction of the feature's impact on the model's prediction: right for an increase and left for a decrease in the likelihood of the positive class. A feature's consistent influence on prediction is suggested by dots clustering on one side, while the spread of dots across the horizontal axis indicates the variability of the feature's impact. The dispersion of dots within a row suggests the presence of interactions with other features that affect the predictive outcome. The ordering of the rows reflects the aggregate strength of the SHAP values across all data points, thus providing a ranked overview of feature importance in the model's predictions.

In the context of minor injuries, our analysis reveals that 'Light' and 'Intersection' significantly influence model predictions, as indicated by their notable SHAP values (Figure 4a). This observation corroborates findings from previous research, which underscores the critical role of lighting conditions and intersection presence in pedestrian crash severity (Harris et al., 2023). Specifically, well-lit areas and intersections, by enhancing visibility and imposing traffic controls, are associated with reduced injury severity following crashes. Such environments provide cues for safer pedestrian and vehicle interaction, potentially mitigating the impact of accidents. Furthermore, the 'VerticalAlignment' factor's prominence in our model aligns with broader safety studies, albeit less directly examined in pedestrian contexts. Insights from research on vehicle dynamics on mountainous freeways suggest that level road geometries contribute to lower crash severities, a principle extendable to pedestrian safety scenarios. Our findings suggest that flat terrains, enhancing pedestrian visibility and influencing vehicle speed, can similarly reduce the severity of pedestrian injuries.

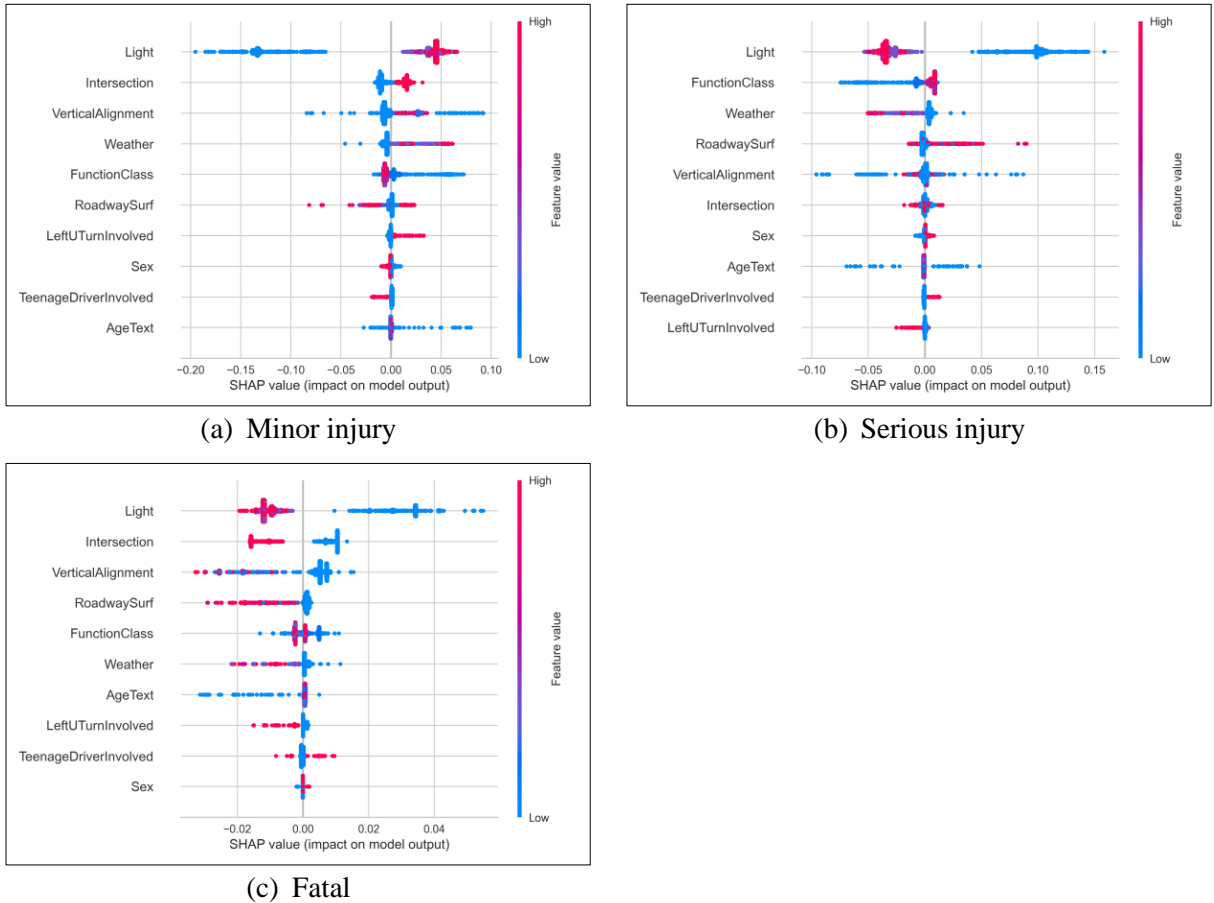

(a) Minor injury

(b) Serious injury

(c) Fatal

**Figure 4. The SHAP summary plot for each crash severity class in Random Forest model**

In our analysis of serious injuries, as depicted in Figure 4b, the 'RoadwaySurf' feature underscores the heightened risk associated with non-dry road conditions, corroborating findings from recent studies that emphasize the adverse effects of wet or icy surfaces on crash severity. These conditions notably compromise vehicle control and extend stopping distances, thereby elevating the likelihood of severe pedestrian injuries. This aligns with research highlighting the critical role of road surface conditions in influencing crash outcomes, where slippery roads are identified as a significant contributor to crash incidence and severity (Lee et al., 2023).

Furthermore, the analysis reveals a critical relationship between lighting conditions and injury severity, with 'Light' indicating that diminished visibility in lower light conditions amplifies the risk of serious injuries. This observation is supported by studies identifying darkness as a key factor in exacerbating pedestrian crash severity, reflecting the challenges drivers face in detecting pedestrians under such conditions (Lee et al., 2023).

Also, the 'FunctionClass' feature demonstrates a differentiated impact on the likelihood of serious injuries in pedestrian crashes. The SHAP values suggest that roads classified as 'Arterial' have a greater association with serious injury outcomes compared to 'Local' or 'Collector' roads. This could be due to arterial roads typically having higher traffic volumes, speeds, and more complex traffic patterns, which may increase the risk and potential severity of pedestrian accidents.

The 'FunctionClass' variable offers additional insights into the risk associated with different road types. Arterial roads, characterized by higher traffic volumes and speeds, are shown to have a stronger association with serious injury outcomes, a finding that echoes research on the impact of road classification on pedestrian safety. Studies from the UK (Salehian et al., 2023) suggest that major roads, analogous to the 'Arterial' classification in our study, pose increased risks due to their traffic characteristics, reinforcing the need for targeted safety measures on such thoroughfares. The convergence of these insights underscores the critical need for enhanced safety protocols and infrastructure improvements on arterial roads to mitigate the higher risks of serious injuries in pedestrian crashes.

In Figure 4c, pertaining to the fatal injury class, the 'Light' feature reveals a profound impact, with darker conditions, particularly 'Dark-Not lighted', significantly increasing the probability of a fatal outcome. This aligns with the expectation that poor lighting can severely impair a driver's visibility and reaction time, leading to more severe accidents (Ferenchak et al., 2022). 'Intersection' also stands out as a critical factor, indicating that incidents occurring at intersections are more likely to result in fatal injuries. This may be due to the complex traffic dynamics and potential for higher-speed collisions at these locations (Siddiqui et al., 2006). Lastly, 'AgeText' appears to have a noticeable effect on the model's predictions for fatal injuries. This suggests that age, as a proxy for factors like physical vulnerability and reaction time, is a significant predictor of the severity of outcomes when pedestrians are involved in traffic incidents, with certain age groups possibly being at a higher risk for fatal outcomes(Kitali et al., 2017).

To navigate the intricacies of the dot summary plot and deepen our understanding of how each EV contributes to the model's predictions, we employed SHAP force plots. Figure 5 showcases such a plot for a specific case in our dataset, observation #313. This visualization (Figure 5a) elucidates the Random Forest model's inclination to categorize the observation as a minor injury, where the prediction score f(x) is 0.82 against a base value of approximately 0.775. Notably, the lighting condition emerges as the most influential factor in this classification.

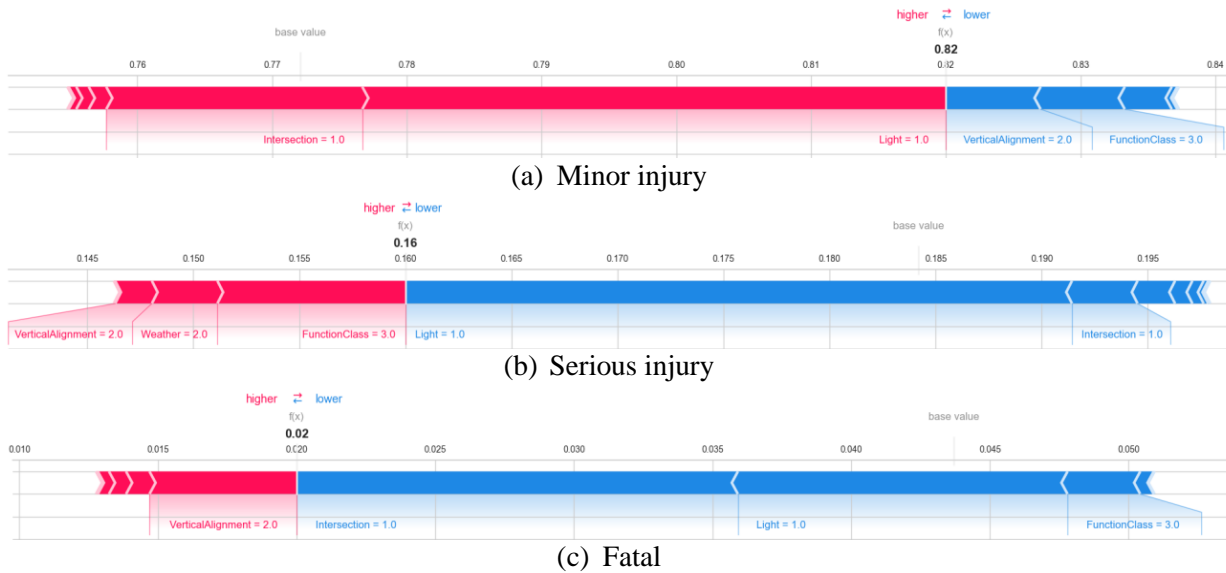

(a) Minor injury

(b) Serious injury

(c) Fatal

**Figure 5. The SHAP values, explaining the contribution of EVs to the raw Random Forest model output for a specific observation.**

## CONCLUSION

In summary, this paper has highlighted the viability of AutoML for pedestrian crash severity analysis, offering a practical alternative to more complex deep learning models such as TabNet, which, although previously applied to this data achieving around 95% accuracy (Rafe & Singleton, 2023b), require greater computational resources and technical knowledge. The relative simplicity and reduced resource needs of AutoML position it as an accessible and efficient solution for diverse applications. We've also developed a user-friendly web application that enables users to build crash severity prediction models via a straightforward data upload process, with AutoML at its core. This tool, along with its source code, has been made available on GitHub (Rafe & Singleton, 2023a), facilitating easy access and collaborative enhancement. Through this research and its outcomes, we encourage further exploration into the integration of user-friendly tools like AutoML in various domains, promoting the adoption of data-driven decision-making in fields where expertise is traditionally siloed.

## REFERENCES

Al-Mistarehi, B. W., Alomari, A. H., Imam, R., & Mashaqba, M. (2022). Using Machine Learning Models to Forecast Severity Level of Traffic Crashes by R Studio and ArcGIS. *Frontiers in Built Environment*, *8*, 860805. https://doi.org/10.3389/fbuil.2022.860805

Angarita-Zapata, J. S., Maestre-Gongora, G., & Calderín, J. F. (2021). A bibliometric analysis and benchmark of machine learning and automl in crash severity prediction: The case study of three colombian cities. *Sensors*, *21*(24). https://doi.org/10.3390/s21248401

Batista, G. E. A. P. A., Bazzan, A. L. C., & Monard, M. C. (2003). Balancing Training Data for Automated Annotation of Keywords: a Case Study. *In Proceedings of the Second Brazilian Workshop on Bioinformatics*, *January*.

Effati, M., & Vahedi Saheli, M. (2022). Examining the influence of rural land uses and accessibility-related factors to estimate pedestrian safety: The use of GIS and machine learning techniques. *International Journal of Transportation Science and Technology*, *11*(1), 144–157. https://doi.org/10.1016/j.ijtst.2021.03.005

Ferenchak, N. N., Gutierrez, R. E., & Singleton, P. A. (2022). Shedding light on the pedestrian safety crisis: An analysis across the injury severity spectrum by lighting condition. *Traffic Injury Prevention*, *23*(7). https://doi.org/10.1080/15389588.2022.2100362

Goswamy, A., Abdel-Aty, M., & Islam, Z. (2023). Factors affecting injury severity at pedestrian crossing locations with Rectangular RAPID Flashing Beacons (RRFB) using XGBoost and random parameters discrete outcome models. *Accident Analysis & Prevention*, *181*, 106937. https://doi.org/10.1016/J.AAP.2022.106937

Harris, L., Ahmad, N., Khattak, A., & Chakraborty, S. (2023). Exploring the Effect of Visibility Factors on Vehicle–Pedestrian Crash Injury Severity. *Https://Doi.Org/10.1177/03611981231164070*, 036119812311640. https://doi.org/10.1177/03611981231164070

Infante, P., Jacinto, G., Afonso, A., Rego, L., Nogueira, V., Quaresma, P., Saias, J., Santos, D., Nogueira, P., Silva, M., Costa, R. P., Gois, P., & Manuel, P. R. (2022). Comparison of Statistical and Machine-Learning Models on Road Traffic Accident Severity Classification. *Computers*, *11*(5), 80. https://doi.org/10.3390/computers11050080

Kang, Y., & Khattak, A. J. (2022). Deep Learning Model for Crash Injury Severity Analysis Using Shapley Additive Explanation Values. *Transportation Research Record: Journal of the*

*Transportation Research Board*, *2676*(12), 242–254. https://doi.org/10.1177/03611981221095087

Kitali, A. E., Kidando, E., Sando, T., Moses, R., & Ozguven, E. E. (2017). Evaluating Aging Pedestrian Crash Severity with Bayesian Complementary Log–Log Model for Improved Prediction Accuracy. *Transportation Research Record*, *2659*(1). https://doi.org/10.3141/2659-17

Komol, M. M. R., Hasan, M. M., Elhenawy, M., Yasmin, S., Masoud, M., & Rakotonirainy, A. (2021). Crash severity analysis of vulnerable road users using machine learning. *PLOS ONE*, *16*(8), e0255828. https://doi.org/10.1371/journal.pone.0255828

Lee, D., Guldmann, J. M., & von Rabenau, B. (2023). Impact of Driver's Age and Gender, Built Environment, and Road Conditions on Crash Severity: A Logit Modeling Approach. *International Journal of Environmental Research and Public Health*, *20*(3). https://doi.org/10.3390/ijerph20032338

Lundberg, S. M., Erion, G. G., & Lee, S.-I. (2018). Consistent individualized feature attribution for tree ensembles. *ArXiv Preprint ArXiv:1802.03888*.

Lundberg, S. M., & Lee, S. I. (2017). A unified approach to interpreting model predictions. *Advances in Neural Information Processing Systems*, *2017-December*.

Moez Ali. (2020). *PyCaret: An open source, low-code machine learning library in Python*. https://pycaret.org/

NHTSA. (2022). *Pedestrian Safety: Prevent Pedestrian Crashes*. https://www.nhtsa.gov/road-safety/pedestrian-safety

Rafe, A., & Singleton, P. A. (2023a). *CrashAutoML*. https://github.com/pozapas/CrashAutoML

Rafe, A., & Singleton, P. A. (2023b). *Exploring Factors Affecting Pedestrian Crash Severity Using TabNet: A Deep Learning Approach*. http://arxiv.org/abs/2312.00066

Rahman, R., Vahedi Saheli, M., & Singleton, P. A. (2023). Risk Factors for Pedestrian Crashes on Utah State Highway Segments: Results from Parametric and Non-Parametric Approaches Controlling for Pedestrian Exposure. *Transportation Research Board (TRB) 102nd Annual Meeting*.

Salehian, A., Aghabayk, K., Seyfi, M. A., & Shiwakoti, N. (2023). Comparative analysis of pedestrian crash severity at United Kingdom rural road intersections and Non-Intersections using latent class clustering and ordered probit model. *Accident Analysis and Prevention*, *192*. https://doi.org/10.1016/j.aap.2023.107231

Shrinivas, V., Bastien, C., Davies, H., Daneshkhah, A., & Hardwicke, J. (2023). Parameters influencing pedestrian injury and severity – A systematic review and meta-analysis. *Transportation Engineering*, *11*, 100158. https://doi.org/10.1016/J.TRENG.2022.100158

Siddiqui, N. A., Chu, X., & Guttenplan, M. (2006). Crossing Locations, Light Conditions, and Pedestrian Injury Severity. *Transportation Research Record: Journal of the Transportation Research Board*, *1982*(1). https://doi.org/10.1177/0361198106198200118

UDPS. (2023). *Utah Crash Summary*. https://udps.numetric.net/utah-crash-summary#/

WHO. (2022). *Road traffic injuries*. https://www.who.int/news-room/fact-sheets/detail/road-traffic-injuries

Yang, L., Aghaabbasi, M., Ali, M., Jan, A., Bouallegue, B., Javed, M. F., & Salem, N. M. (2022). Comparative Analysis of the Optimized KNN, SVM, and Ensemble DT Models Using Bayesian Optimization for Predicting Pedestrian Fatalities: An Advance towards Realizing the Sustainable Safety of Pedestrians. *Sustainability*, *14*(17), 10467. https://doi.org/10.3390/su141710467